\documentclass[runningheads]{llncs}
\usepackage{eccv}
\usepackage{eccvabbrv}

\usepackage{graphicx}
\usepackage{booktabs}
\usepackage{caption}
\usepackage{subcaption}
\usepackage{booktabs}
\usepackage{siunitx}
\usepackage{amsmath}
\usepackage{multirow} 
\usepackage{tabularx} 
\usepackage{lipsum} 
\usepackage{booktabs} 
\usepackage{array} 
\usepackage[dvipsnames]{xcolor}
\usepackage[accsupp]{axessibility}  
\usepackage{arydshln}

\usepackage{hyperref}

\usepackage{orcidlink}
\usepackage{xcolor} %
\usepackage{adjustbox}
\usepackage{wrapfig}
\usepackage{graphicx} 
\usepackage{algorithm}
\usepackage{algpseudocode}
\usepackage{amsmath}
\usepackage{xcolor}

\newcommand{\crule}[3][black]{\textcolor{#1}{\rule{#2}{#3}}}

\begin{document}
 
\title{They're All Doctors: Synthesizing Diverse Counterfactuals to Mitigate Associative Bias} 
 
\author{Salma Abdel Magid\inst{1}\and
Jui-Hsien Wang\inst{2} \and
Kushal Kafle\inst{2}\and Hanspeter Pfister\inst{1}}
 
\authorrunning{S.~Abdel Magid et al.}
 \titlerunning{Preprint}
\institute{Harvard University \and
Adobe Research\\
\email{\{sabdelmagid,pfister\}@g.harvard.edu} \\
\email{\{jwang,kkafle\}@adobe.com}
}

\maketitle

\begin{abstract}
 
    Vision Language Models (VLMs) such as CLIP are powerful models; however they can exhibit unwanted biases, making them less safe when deployed directly in applications such as text-to-image, text-to-video retrievals, reverse search, or classification tasks. In this work, we propose a novel framework to generate synthetic counterfactual images to create a diverse and balanced dataset that can be used to fine-tune CLIP. Given a set of diverse synthetic base images from text-to-image models, we leverage off-the-shelf segmentation and inpainting models to place humans with diverse visual appearances \emph{in context}. We show that CLIP trained on such datasets learns to disentangle the human appearance from the context of an image, i.e., what makes a doctor is not correlated to the person's visual appearance, like skin color or body type, but to the context, such as background, the attire they are wearing, or the objects they are holding. We demonstrate that our fine-tuned CLIP model, $CF_\alpha$, improves key fairness metrics such as MaxSkew, MinSkew, and NDKL by 40-66\% for image retrieval tasks, while still achieving similar levels of performance in downstream tasks. We show that, by design, our model retains maximal compatibility with the original CLIP models, and can be easily controlled to support different accuracy versus fairness trade-offs in a plug-n-play fashion.

      \end{abstract}
 
\section{Introduction}
\label{sec:intro}

\begin{figure}[ht]
    \centering
    \includegraphics[width=1\linewidth]{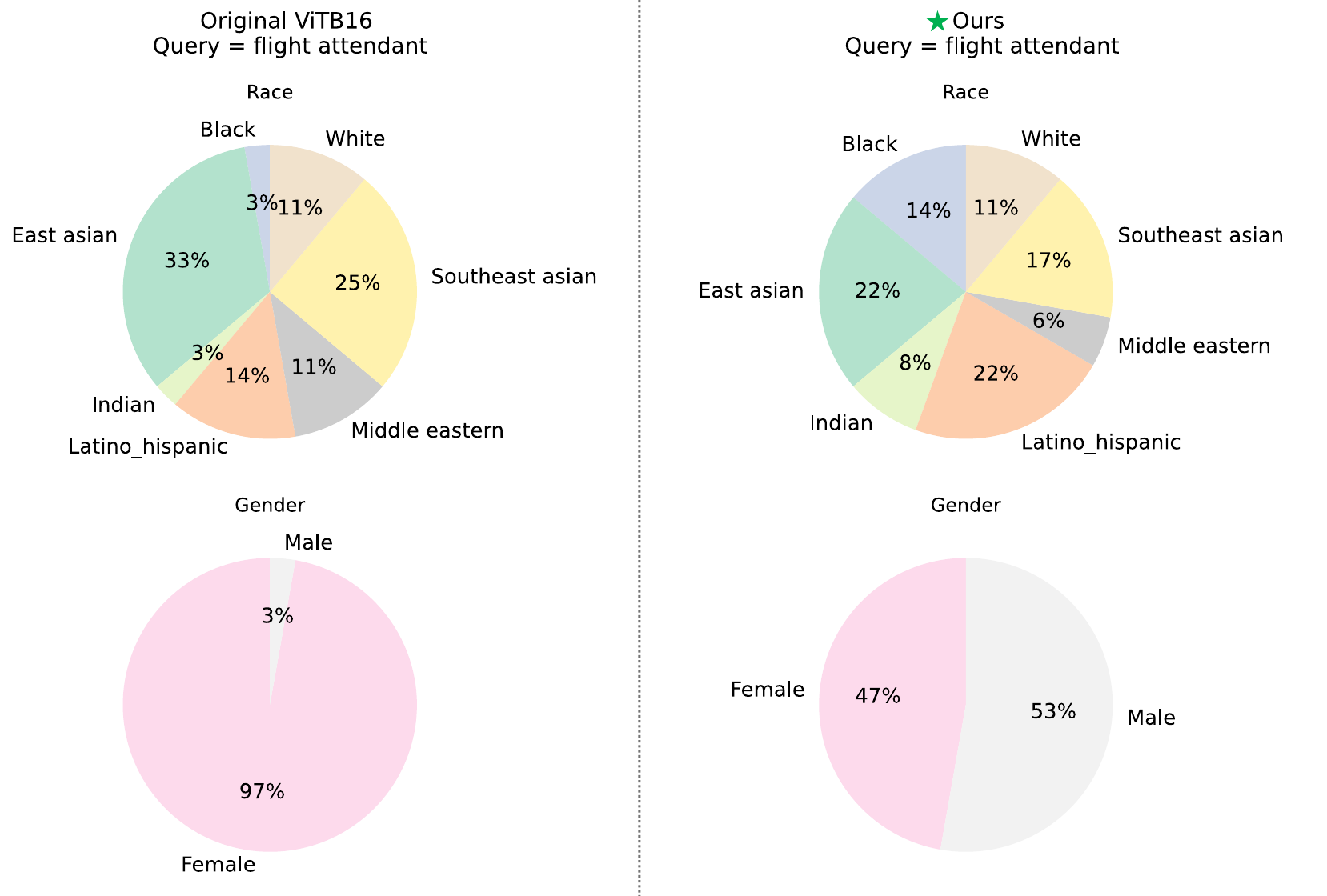}
    \caption{After training on our counterfactual image dataset, a fine-tuned CLIP ViT-B/16 model can retrieve more uniform image distributions for different races and gender for query "flight attendent" on the FairFace dataset. }
    \label{fig:topkteaser}
\end{figure}
 
Large vision language models (VLMs) pretrained on very large datasets of image and text pairs show impressive performance on various tasks~\cite{pmlr-v139-radford21a,wortsman2022robust,baldrati2022effective, saito2023pic2word,couairon2022embedding,luo2022clip4clip,pei2023clipping,unumcloud_2023_efficient,lin2023gridclip,kirillov2023segment,mokady2021clipcap}. However, this discriminative capability comes with a catch: as shown in previous work, large datasets used to train VLMs, such as the LAION dataset~\cite{schuhmann2021laion,schuhmann2022laion}, can 
contain harmful content and various biases which are then transferred to the learned model~\cite{birhane2024into,bender2021dangers}, which in turn leads to biased retrieval results, wrongful classification/characterization, and other unwanted behaviors. 
 
This problem is especially notorious in the context of human-related retrieval queries. For example, CLIP strongly associates the profession "flight attendant" with images of women, even if the source image set contains only tightly cropped faces (see Figure~\ref{fig:topkteaser}).

One straightforward way to address these types of issues is by manipulating the training dataset. However,  doing so in a fully supervised way is challenging. To illustrate this, we first consider the following \emph{desiderata} of an ideal dataset:
\begin{itemize}
    \item \textbf{Size}: the dataset size should be large enough to support training of a transformer model such as those used in the CLIP ViT model family.
    \item \textbf{Subject}: since we are primarily concerned about model performance on human-related queries, the dataset will be centered around humans. Specifically, subjects should be clearly visible in each of the images.
    \item \textbf{Diversity}: images in the dataset should be diverse in the visual concepts that we are trying to debias. For example, having diverse images of doctors in different scenes and contexts enables the model to learn stronger representations of doctors.  
    \item \textbf{Balance}: on the other hand, images should be balanced against protected attributes and appearances. In addition to the different contexts and scenes, "doctors" should ideally appear with different skin colors, hair styles, gender expressions, body types, etc.

\end{itemize}
Posed like this, we can clearly see the challenges of collecting such a dataset: 1) it is difficult to find images while respecting diversity and balance at scale. That is, for each protected intersectional group, we must have a sufficiently large number of images. 2) It is crucial to ensure that the joint distribution of nonprotected attributes like context, lighting, pose, etc. is equal across protected attribute groups. However, different cultures and social norms might associate the same concept to different visual appearances. For example, a doctor's attire or visual props might look one way in one cultural context, but look very different in another cultural context. 3) Collecting and labeling such a dataset requires significant resources, can be susceptible to other sets of biases such as annotator bias~\cite{geva2019we}, and raises issues of privacy. Thus, it is near impossible to do this with data that we cannot construct or intervene on ourselves. 

\begin{figure}[tb]
  \centering
  \includegraphics[width=\columnwidth]{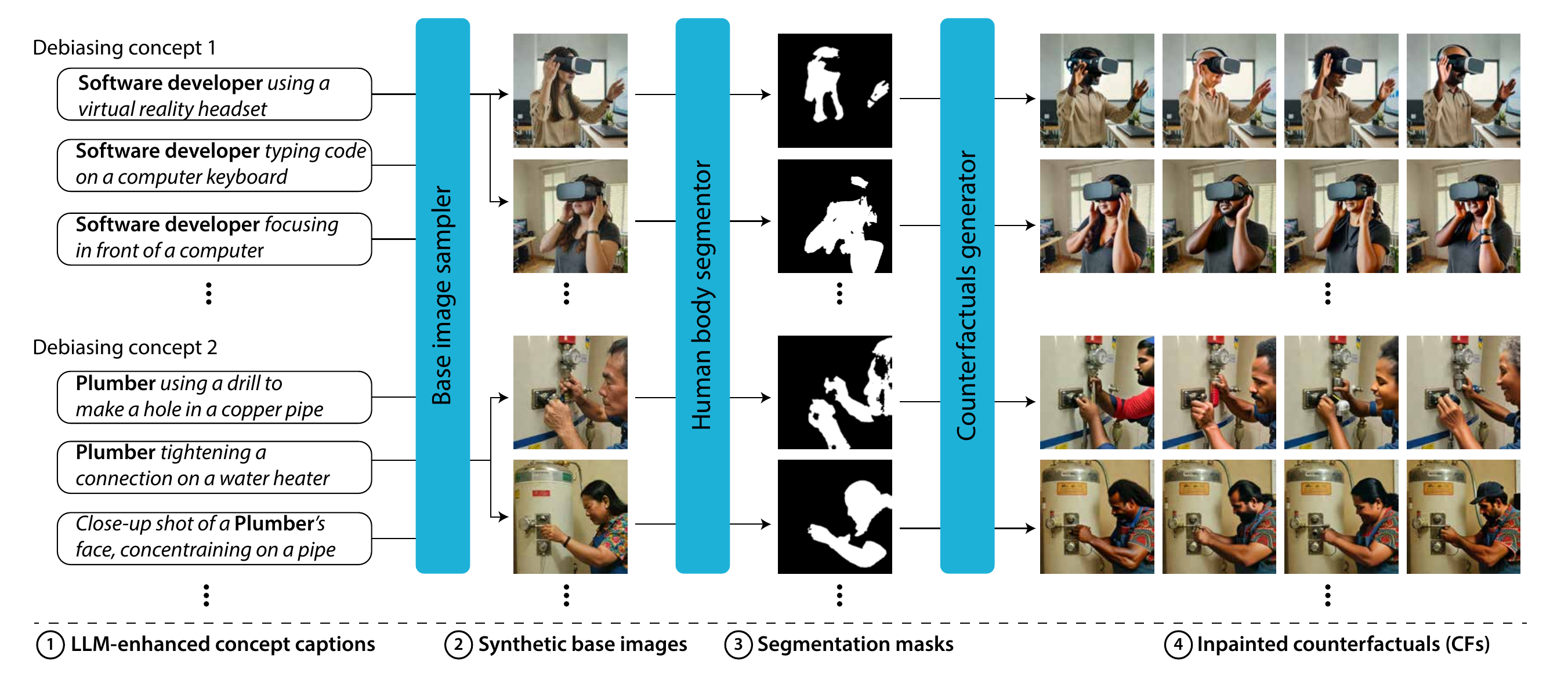}
  \caption{\textbf{Synthetic counterfactual generation overview} For each base textual keyword such as the profession "software developer", we first use LLMs to generate a set of plausible captions. Each caption then gets sampled with additional decorator inputs to generate a set of base images. We then compute masks that correspond to human body parts, and finally inpaint with decorators to synthesize counterfactuals. This pipeline enables us to generate a large amount of counterfactuals with diverse humans, while controlling visual cues that might cause secondary association for the downstream training tasks, such as the background or a prop a subject is holding.  
  }
  \label{fig:overview}
\end{figure}

To address these challenges, we introduce a method that generates minimal-change counterfactuals through masking to reduce associative bias in CLIP (see Figure~\ref{fig:overview}). These counterfactuals are then used as labels from the same class so that we can train CLIP with a loss that encourages image representations of counterfactuals to be close together in the embedding space. Our experiments show that fine-tuning CLIP with our synthetically augmented dataset -- comprising of entirely synthetic captions and images -- significantly improve its fairness when evaluated on real images.  We also introduce a simple, interpretable technique to control the fairness and accuracy trade-off using  weight interpolation techniques, so that the user can determine performance on downstream tasks. Our approach is motivated by the understanding that a doctor's appearance is defined by their attire, background, and props, rather than by skin tone, perceived gender, age, or other sensitive attributes. By masking and inpainting these image parts, we direct the model to focus on contextual cues, thus eliminating spurious correlations learned from imbalanced datasets. Our framework operates fully automatically, utilizing synthetic data (images and captions) throughout, ensuring privacy and generating safe captions, as it is completely controllable and synthetic.  \\

Our core contributions are:

\begin{enumerate}
  \item We present a practical framework for creating diverse and balanced image datasets starting from a set of core visual concepts to debias, such as professions.
  \item We integrate two techniques for fine-tuning CLIP, linearly blending weights and an additional self-supervised loss term which explicitly uses the counterfactuals. 
  \item We demonstrate that CLIP's bias on sensitive attributes, such as race and gender, can be mitigated by finetuning using our dataset out-of-the-box. We also show that our model is fully compatible with the original pretrained model. This enables users to control the accuracy-fairness tradeoff.
\end{enumerate}

\paragraph{Scope for visual concepts.}
Although CLIP exhibits different levels of biases towards various types of textual inputs (negative or positive connoted words, professions, etc.), we strategically focus on debiasing professions for this paper. This is because simpler baseline strategies can be applied in the cases of non-profession inputs: one can neutralize negatively connoted words and filter out slurs (e.g., modify "stupid person" query to just "person"). However, there is not an effective solution towards professions. To make matters worse, these search terms are usually specified in an attribute-neutral way (e.g., a user typically searches for "pilot" rather than "Asian male pilot"). Consistent with the findings in the seminal BIG Bench paper~\cite{srivastava2023bigbench}, we find that broad and under-specified profession-based queries often result in heavily biased results. We therefore intentionally focus on debiasing professions, while keeping our framework flexible enough to handle a broader case of concepts.

\section{Related Work}

\paragraph{Vision Language Models and biases.}
Large VLMs have garnered a lot of attention over the past few years due to their capability to learn joint information between text and image from a large amount of data~\cite{pmlr-v139-radford21a,li2022blip,lu2019vilbert,kim2021vilt,jia2021scaling,zhang2021vinvl,singh2022flava}, and transfer their knowledge to downstream computer vision tasks~\cite{pmlr-v139-radford21a,wortsman2022robust,baldrati2022effective, saito2023pic2word,couairon2022embedding,luo2022clip4clip,pei2023clipping,unumcloud_2023_efficient,lin2023gridclip,kirillov2023segment,mokady2021clipcap}. Here we focus on CLIP~\cite{pmlr-v139-radford21a}  due to its impressive performance and widespread applications. Many recent works have documented the bias issues in CLIP, such as performance gaps between perceived gender \cite{hall2023vision} on various downstream applications.  Various evaluation metrics for measuring bias \cite{berg2022prompt} and attempts at defining a taxonomy for bias evaluation \cite{ali2023evaluating} have also been explored. 

\paragraph{Bias mitigation.}
Techniques for mitigating bias in VLMs can broadly be categorized based on which stage the intervention occurs: pre-processing, in-processing (during training), and post-processing, or some combination of these. Techniques such as~\cite{alabdulmohsin2023clip,wang2023overwriting} aim to augment the training data, either by re-sampling or re-weighting data points to satisfy fairness criteria. In-processing debiasing techniques include making adjustments to the model, objective, or overall strategy for joining modalities~\cite{berg2022prompt,seth2023dear}. Post-processing techniques freeze the model and operate in the embedding space. \cite{chuang2023debiasing,tanjim2024wacv} showed that biases can be reduced by projecting out certain subspaces in the text embedding. \cite{wang2021gender} demonstrated that feature clipping can reduce gender bias in image search. Our method is a combination of pre-processsing and in-processing as we not only create a new fair training dataset, but also finetune the model with adjustments to the loss to explicitly make use of the counterfactuals during training.

\paragraph{Synthetic data for fairness.}
The use of synthetic data to evaluate the fairness of VLMs has also been explored. PAIRS~\cite{fraser2024examining} examines racial and gender biases by generating synthetic images. GenSynth~\cite{smith2023balancing} edits genders in the COCO dataset~\cite{lin2014microsoft}. Aimed primarily for evaluation, the scale of the data in these works is too small to be used for training large VLMs. Other works have also explored utilizing counterfactuals for evaluation, specifically to examine how a classifier's prediction can change if the input face changes slightly~\cite{denton2019image,zhang2022counterfactually,zhao2021understanding}.  
\cite{luccioni2024stable} audits the social biases embedded in text2image models, by comparing model outputs for a targeted set of prompts. \cite{esposito2023mitigating} leveraged image generation to fine-tune text-to-image models. \cite{friedrich2023fair} proposed a strategy for bias mitigation at the deployment stage of diffusion models. Our work on creating a large-scale, diverse dataset that can be used to fine-tune CLIP is complementary to this line of work.

\section{Background}

\subsection{Ethical Considerations}
We must acknowledge that sensitive attributes such as gender, race, and nationality are social constructs that can not be captured through binary or fixed categories~\cite{hanna2020towards,hu2020s}. These attributes have salient material effects, are multi-dimensional, and must be located along the spectra by the individuals themselves. By only using reductive categories, we may be perpetuating algorithmic unfairness \cite{luccioni2024stable}.  However, we explicitly design our framework such that the model being fine-tuned is never exposed to any gendered or racial language. The descriptors are only used for generating diverse counterfactuals, recognizing that our approach operates within the limitations of stereotype-based generation mechanisms inherent to current text2image (T2I) models. We include a longer discussion of the ethical implications of this work in the supplementary, and detail issues of stacking biases (whether from LLMs or T2I models) in the subsequent sections.

\subsection{VLM and constrastive learning}

Contrastive Language-Image Pre-training (CLIP) is a popular form of VLM that was proposed in~\cite{pmlr-v139-radford21a}. CLIP contains an image encoder, $\mathcal{I}$, and a text encoder, $\mathcal{T}$. Given a text-image input pair $(\boldsymbol{x}_t, \boldsymbol{x}_i)$, the contrastive loss in CLIP conceptually brings $\mathcal{I}(\boldsymbol{x}_i)$ close to $\mathcal{T}(\boldsymbol{x}_t)$ while maximizing the distance to other pairs in the same batch, resulting in a joint embedding space for image and text. Measures in this co-embedded space such as the simple cosine similarity between a new image and text pair, $\mathcal{D}(\mathcal{I}(\boldsymbol{y}_i), \mathcal{T}(\boldsymbol{y}_t))$, can then be used to quantify their similarity, leading to applications such as free-form text-to-image retrieval.

\section{Methodology}

\subsection{Counterfactual Image Generation Algorithm}
\label{sec:counterfactual_generation}

The key insight to our algorithm is to leverage the precise control of generative models to place visually diverse humans in the same context. This will encourage the model being debiased to learn the context associated to a concept (e.g. "doctor"), rather than through sensitive attributes of humans. To do this, we utilize human body masking followed by inpainting to replace subjects of interest (see Figure~\ref{fig:overview}). Our algorithm contains four key steps (Algorithm~\ref{alg:enhanced_CLIP_pre_generated}), each of which will be outlined below.

\begin{algorithm}
\caption{Generating Diverse Counterfactuals}\label{alg:enhanced_CLIP_pre_generated}
\begin{algorithmic}[1]
\State \textbf{Input:} Set of debiasing concepts $\mathcal{P}$, decorator sampler $\mathcal{S}$ for diverse attributes (see Table~\ref{table:decorator_prompts}), Large-Language Model (LLM).
\State \textbf{Output:} Set of counterfactual images $\mathcal{C}$.
\State $\mathcal{C} \gets \emptyset$
\For{each concept $p \in \mathcal{P}$}
    \State $\boldsymbol{C}_p \gets$ \texttt{GenerateCaptions}($p$, LLM)
    \State $\mathcal{B}_p \gets \emptyset$
    \For{each caption $\boldsymbol{c} \in \boldsymbol{C}_p$}
        \State $\mathcal{B}_p \gets \mathcal{B}_p \cup $ \texttt{SampleBaseImages}($\boldsymbol{c}$, $\mathcal{S}$)
    \EndFor
    \For{each base image $b \in \mathcal{B}_p$}
        \State  $\mathcal{M} \gets$ \texttt{SegmentHumanBodyParts}($b$)  \Comment{Compute inpainting masks}
        \State $\mathcal{C} \gets \mathcal{C} \cup \texttt{GenerateCounterfactuals}(b, \mathcal{M}, \mathcal{S})$
    \EndFor
\EndFor

\end{algorithmic}
\end{algorithm}
 
\noindent\paragraph{Generating neutral captions.}
\label{sec:step1basecaptions}

\begin{table}[ht]
\centering
\caption{Sample generated captions for various professions.}
\label{table:sample_captions}
\begin{tabular}{ll}
\hline
\textbf{Profession} & \textbf{Sample Caption} \\
\hline
Architect & "An architect reviewing blueprints in a bright office." \\
Chef & "A chef garnishing a dish in a professional kitchen setup." \\
Photographer & "A photographer adjusting the lens of a DSLR camera outdoors." \\
Teacher & "A teacher explaining a concept on a digital whiteboard in a classroom." \\
Nurse & "A nurse preparing medication in a hospital's patient room." \\
\hline
\end{tabular}
\end{table}

For each visual concept $p$ to be debiased, we utilize off-the-shelf LLM model LLaMA 70B~\cite{touvron2023llama} to prompt for a set of $n_p$ text captions. Because we are focusing on professions (see \S\ref{sec:intro} for details), the prompting strategy includes detailing the concrete equipment, tasks, and environment characteristic of each profession, similar to the strategy used in previous work~\cite{fan2024improving,menon2022visual,pratt2023does}  This allows the downstream text-to-image model to be able to adhere to the caption to generate an image of the profession featuring a single individual. The prompts and output captions can be found in the supplemental material.

Critically, a requirement for our generated captions is that they must be neutral with respect to all protected attributes. For example, "a man pilot" will be rejected as it contains the "man" attribute. This is for two purposes: 1) to avoid associative biases that can arise from LLM's own training data, and 2) the downstream constrastive training is done with the attribute-neutral captions to target biases stemmed from under-specification. In addition, we battle LLM's known prompt adherence issues and hallucination by a small set of natural language processing routines (i.e., removing gender and racial language and gendered pronouns). We provide sample captions in Table~\ref{table:sample_captions}, and will release the dataset containing all the generated captions for reproducibility.

\noindent\paragraph{Generating base images.}
\label{sec:step2baseimages}
Recall that the desiderata of our data generation is to have balance and diversity. Directly using the neutralized captions alone will not synthesize diverse base images. Our goal is to not only synthesize diverse faces (in the counterfactual step), but to ensure that \textit{the base images which are being inpainted into are also diverse}. This can be controlled with various decorators about the scene and the subject. We thus generate multiple base images, each with different decorators, for each caption. The design of our decorator categories follows previous work~\cite{esposito2023mitigating}, and is shown in Table~\ref{table:decorator_prompts}.  This allows us to generate sets of visually rich images for each caption (see Figure~\ref{fig:overview}). We use SDXL Turbo~\cite{sauer2023adversarial} for the generation, a fast diffusion model that can generate high fidelity images in a \textit{single} diffusion step, making our method computationally efficient.

\begin{table}[ht]
\centering
\captionsetup{font={scriptsize}} 
\caption{Sample decorators for image generation (see \S\ref{sec:step2baseimages} for details). The full list can be found in the supplementary.}
\label{table:decorator_prompts}
\begin{adjustbox}{width=0.65\columnwidth,totalheight=\textheight,keepaspectratio,center}
\begin{tabular}{ll}
\hline
\textbf{Attribute} & \textbf{Examples} \\
\hline
Shot Style & iPhone, Long shot, Upper body shot \\
Nationality & Nigerian, Indian, Vietnamese, Egyptian \\
Age & Young (20s), Middle-aged (30s-40s), Mature (50s+) \\
Gender & Man, Woman \\
Skin Color & Brown, Black, Light \\
Body Type & Fat, Skinny, Chubby, Athletic \\
Hair Color & Black, Brown, Red, Grey \\
Hair Style & Afro, Long, Bald, Ponytail, Braids \\
\hline
\end{tabular}
\end{adjustbox}
\end{table}

\noindent\paragraph{Generating masks for base images.}
For each base image, we perform segmentation to isolate parts that are identifiable for any protected attribute. Since we care about race and gender debiasing, concretely, this equates to parts associated to a human body, such as "hair", "face", "legs", "arms" classes. We do not consider attire and covered body parts to be attribute identifiable, though we acknowledge that attire and pose can be proxies for attributes. The model we utilized is a SegFormer~\cite{segformer} fine-tuned on the ATR dataset~\cite{ATR1}. 

\paragraph{Generating counterfactual images}
Given the base image and the masks, we generate the final counterfactual images by using an off-the-shelf inpainting model, namely, SDXL version 1.0~\cite{podell2023sdxl}. Figure~\ref{fig:overview} shows examples of the generated counterfactual images and their corresponding captions; more can be found in the supplemental material. Our method is in line with the theoretical and empirical findings of the InfoMin principle \cite{tian2020makes}, which states that good counterfactuals for a given task in contrastive representation learning framework should retain task-relevant information while minimizing irrelevant nuisances.

\paragraph{Counteracting biases in generative models.}
Generative T2I models have biases of their own.  We counter this problem with two approaches: 1) consistent with the observation that under-specification increases bias and vice versa~\cite{srivastava2023bigbench}, we observed that by giving clear instructions on the attributes of a specific image, T2I models can more closely adhere to the prompt. For example, the T2I model can show bias towards a specific gender when prompted with just "CEO", but much less bias when prompted with "male CEO" and "female CEO". In other words, the model is less likely to bias in an attribute direction if that attribute is already specified; 2) we use negative prompting with words such as "nudity", "makeup", "jewelry" and a slight adjustment on the decorator distribution to favor minority groups. However, the model can still fail to adhere to the prompts. We illustrate failure cases in the supplementary material. The dataset and datasheet \cite{gebru2021datasheets} will be released.

\subsection{Fine-Tuning CLIP On Counterfactual Data}
\label{sec:fine_tuning_clip}

\begin{figure}[tb]
  \centering
  \includegraphics[width=1.0\columnwidth]{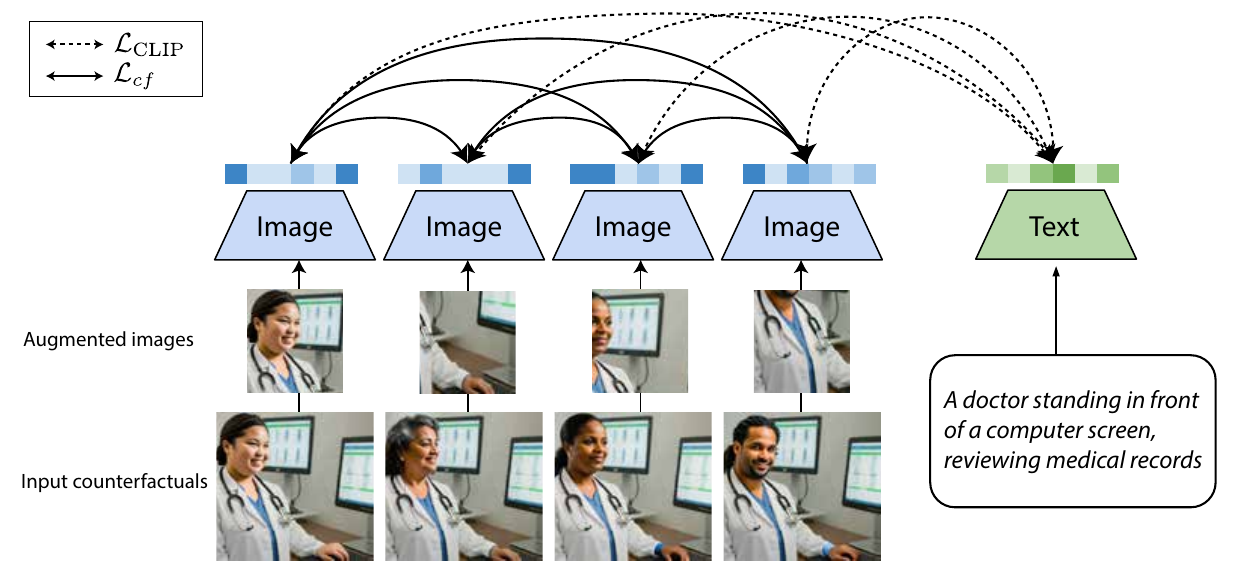}

  \caption{\textbf{Counterfactual loss}, $\mathcal{L}_{cf}$, is added to additionally contrast counterfactual images of the same base image; typical text-to-image and image-to-text contrastive loss, $\mathcal{L}_{\text{CLIP}}$, is also used during our training. Similar to ~\cite{tian2024stablerep} and ~\cite{mu2022slip}. We use random cropping to augment data similar to SimCLR~\cite{chen2020simple}.}

  \label{fig:cf_loss}
\end{figure}

We hypothesize that fine-tuning CLIP on the counterfactual data allows the model to re-learn the representation about the debiasing concepts such that they become disentangled from the protected attributes. To this end, we propose to better leverage our structured data during training. Since all the counterfactual images originating from the same base image share the same visual concept, we want to push them closer in the latent space (see Figure~\ref{fig:cf_loss}). To achieve this, we use an additional constrastive loss term, similar to ~\cite{tian2024stablerep} and ~\cite{mu2022slip}, between each counterfactual image pair, $\boldsymbol{x}_{i}^{cf}$ and $\boldsymbol{x}_j^{cf}$, and their encoded latents, $\boldsymbol{z}_{i} = \mathcal{I}(\boldsymbol{x}_i^{cf})$ and $\boldsymbol{z}_{j} = \mathcal{I}(\boldsymbol{x}_j^{cf})$ (dropping superscript for convenience), within the same batch as
\begin{align}
    \boldsymbol{l}_{i,j} = - \log \frac{\exp(\boldsymbol{z}_{i}^\intercal \boldsymbol{z}_j) / \tau}{\sum_{k \in \text{batch}} \mathbb{1}_{[k \neq i]} \exp(\boldsymbol{z}_i^\intercal \boldsymbol{z}_k) / \tau)}
    \label{eq:counterfactual_loss}
\end{align}
where $\tau \in \mathbb{R}$ is a learnable temperature parameter, and $\mathbb{1}$ is the indicator function.  The final training loss is a combination of the original text-image contrastive loss and the counterfactual loss:
\begin{equation}
\label{eq:lossfunction}
    \mathcal{L} = \beta_1\mathcal{L}_{\text{CLIP}} + \beta_0 \mathcal{L}_{\text{cf}}
\end{equation} 
where $\mathcal{L}_{\text{cf}} = \sum_{(i,j) \in \text{batch}}\boldsymbol{l}_{i,j}$ sums over all positive pairs of counterfactual images in the batch, and $\beta_0, \beta_1 \in \mathbb{R}$ are hyperparameters. By construct, each counterfactual is associated with a base image and caption, and thus no additional labels are required.

\subsection{Controlling The Accuracy-Fairness Trade-Off}
\label{sec:accuracy_fairness_tradeoffs}
Previous work has shown that fine-tuning CLIP can result in catastrophic forgetting~\cite{andreassen2021evolution}, and sensitivity to hyperparameters~\cite{wang2023overwriting}. We intentionally design our framework so that we can leverage the WiSE-FT method~\cite{wortsman2022robust} for weight-space ensembling. To control the accuracy-fairness trade-off, one can simply ensemble the model weights using linear blending $\theta = (1 - \alpha)\theta_{\text{CLIP}} + \alpha\theta_{\text{ours}}$. Effects of this blending are analyzed in the results section~\S\ref{sec:results_alpha_blend}. 

\section{Experiments}

\subsection{Implementation Details}
\label{sec:implementation_details}
Each of our models are fined-tune from different CLIP variants for 30 epochs, utilizing early stopping. The model training follows the loss defined in Equation~\eqref{eq:lossfunction}. We use a learning rate of $1e-5$ with a weight decay of 0.1, and use the AdamW optimizer with $\beta_0=0.9$ and $\beta_1=0.95$. During each training batch, we randomly sample 512 captions from the dataset. For each caption, $m=3$ counterfactuals are then randomly sampled. In a distributed setting, our effective batch size is 1536. Random resized crop at $224\times224$ is always applied. The professions dataset comprises approximately 8,300 unique captions generated from roughly 270 professions aggregated from various sources~\cite{friedrich2023fair,nadeem2020stereoset,luccioni2024stable}, and U.S. Labor Bureau of Statistics data. With 3 base images per caption and 4 counterfactual samples per base image, for a total of approximately 100k images. The non-professions dataset, similarly structured, also results in approximately 100k images.  

\subsection{Evaluation Metrics}
\label{sec:evaluation_metrics}
Following the formulation of previous work~\cite{geyik2019fairness,seth2023dear, chuang2023debiasing}, we quantify bias in our model using standard metrics including
maximum/minimum skew, and the normalized discounted Kullback-Leibler divergence (NDKL). Skew captures the largest unfair advantage or disadvantage across attributes, while NDKL is a retrieval metric that measures divergence from an unbiased distribution. We include the precise definitions in the supplementary. In addition to the fairness metrics, we report accuracy of our model on downstream tasks with FlickrR@5: Recall@5 text-to-image retrieval on the Flickr-1k test set, and ImageNet1KAcc: Image classification accuracy on the ImageNet-1k validation set, following \cite{seth2023dear}.
 
\subsection{Evaluation Datasets}
Although our model is trained on a synthetic dataset, we evaluate it on real-world datasets. Specifically, we use FairFace~\cite{karkkainen2019fairface}, and the Protected Attribute Tag Association (PATA) dataset~\cite{seth2023dear}. FairFace consists of image crops of faces, and their corresponding race, gender, and age annotations collected from Flickr. PATA is a context-based bias benchmarking dataset for evaluating the fairness of large pre-trained VLMs. It consists of annotated images for race, gender, and age for 20 scenes.\footnote{As of February 24, a small part of image URLs in PATA are invalid.} For both datasets, we remove images where the annotated age is below 19 years old and re-balance the data with respect to race and gender.

\subsection{Fairness Evaluation}
Table \ref{tab:bias_metrics} shows consistently improved fairness for our model across transformer variants. For instance, for FairFace, both the ViT B/16 and ViT B/32 models with $CF_{0.5}$ show a significant decrease in MaxSkew@1K and NDKL, achieving a reduction of almost half than the baseline. Similar trends of bias reduction are observed within the PATA dataset. The weight ensembling technique allows us to retain most of the performance on other downstream tasks, shown in Table~\ref{tab:performance_metrics}. 
 
We also compute the average per-group recall for each scene in PATA. Table \ref{tab:pergroup_recall} demonstrates how the recall scores for each group increased, especially for the worst-performing groups, Indian Woman (IW) and Black Man (BM), by 5\% and 4\% respectively. In turn, the disparity in recall scores, which is the difference between the recall for the best performing group and the recall for the worst performing group, also decreased from 8\% to 6\%.

\begin{figure}[tb]
  \centering
  \includegraphics[width=1.0\columnwidth]{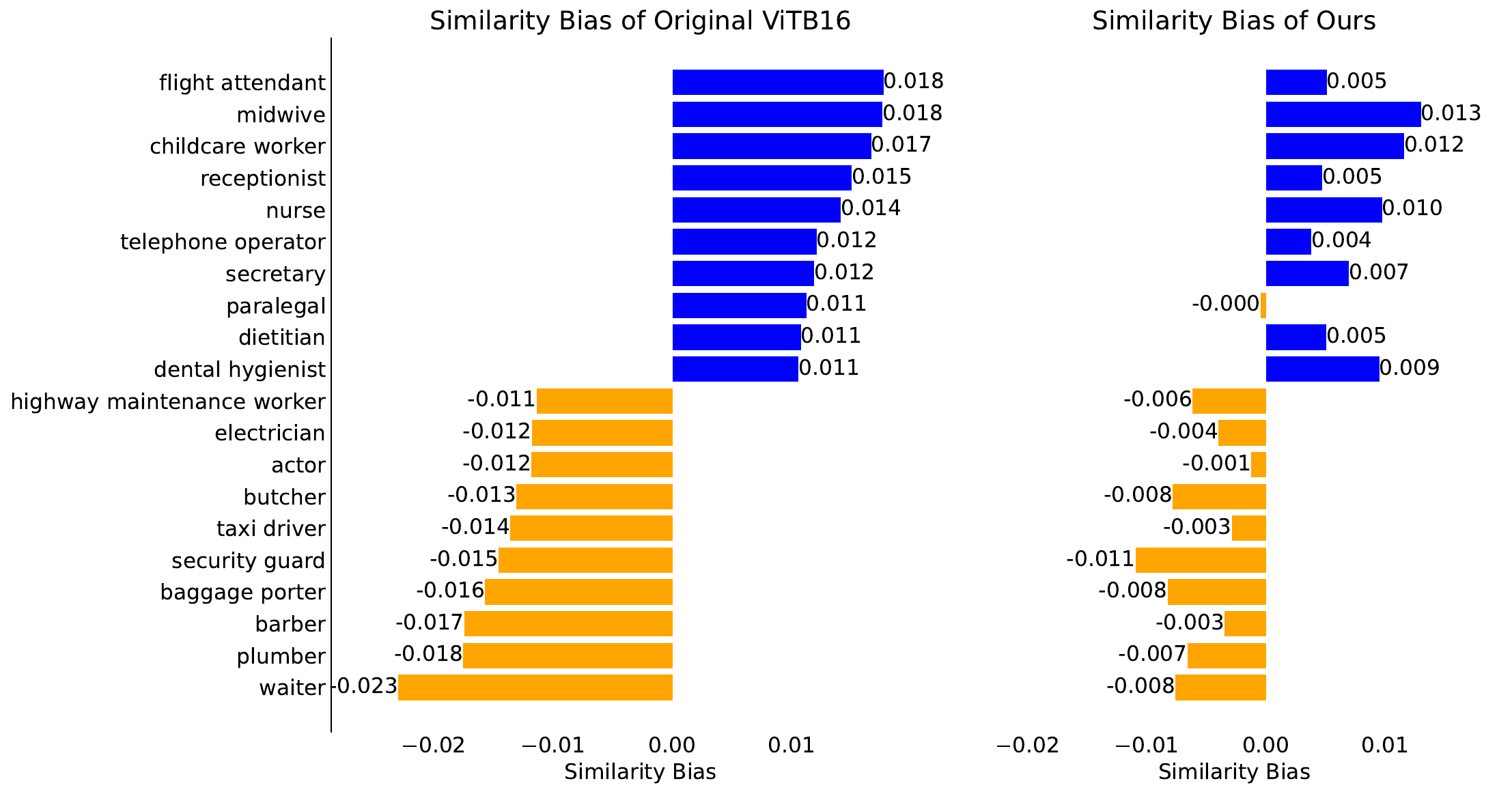}

  \caption{The gender similarity bias measured on PATA. We visualize the similarity biases on the top 20 professions. \crule[orange]{8pt}{8pt} indicates the profession is biased towards men and \crule[Blue]{8pt}{8pt} indicates the profession is biased towards women. Our framework mitigates gender bias for a variety of occupations.}

  \label{fig:similarity_bias_qual}
\end{figure}

Moreoever, in a real-world scenario we may search image corpora that may or may not contain the queried profession. To emulate this, we query PATA with the professions seen during training and show the top 20 most gender biased words along with their similarity bias in Figure \ref{fig:similarity_bias_qual}. We adjust the definition in \cite{wang2021gender} and formulate similarity bias as the difference of expected cosine similarity between images annotated with the \textit{woman} label and images annotated with the \textit{man} label. Given a profession $p$, 
\begin{equation}
    simbias(p) = \mathbb{E}_{v \in \mathcal{V}_W} S(v, p) - \mathbb{E}_{v \in \mathcal{V}_M} S(v, p)
\end{equation}
where $\mathcal{V}_W$ and $\mathcal{V}_M$ are the sets of images labeled as \textit{woman} and \textit{man}. 

Figure \ref{fig:similarity_bias_qual} shows how our method can significantly reduce the similarity bias. The original pre-trained CLIP exhibits large differences in the similarity scores between images of men and images of women, with the largest bias occurring for waiter, plumber, flight attendant, and midwive. For waiter and flight attendant, our method reduces the similarity bias by approximately 3 times, and reduces the similarity bias of actor and paralegal to nearly zero. Overall, our method obtains a 55.5\% lower mean absolute bias of these 20 professions compared to the CLIP model ($0.00645$ versus $0.0145$).

\begin{table*}[t!]
\centering
\footnotesize
\caption{Comparison of bias of models fine-tuned on our synthetic dataset using weight ensembling set at $\alpha=0.5$. R: race; G: gender, RG: race and gender together.}
\setlength{\tabcolsep}{3pt}
\begin{tabular}{@{}lccccccc@{}}
\toprule
\multirow{2}{*}{\textbf{Dataset}} & \multirow{2}{*}{\textbf{Model}} & \multicolumn{6}{c}{\textbf{Bias$\downarrow$}} \\
\cmidrule(l){3-8} 
 & & \multicolumn{3}{c}{\textit{MaxSkew@K}} & \multicolumn{3}{c}{\textit{NDKL}} \\
 & & \multicolumn{1}{c}{R} & \multicolumn{1}{c}{G} & \multicolumn{1}{c}{RG} & \multicolumn{1}{c}{R} & \multicolumn{1}{c}{G} & \multicolumn{1}{c}{RG} \\
\midrule

\multirow{4}{*}{FairFace} 
 & ViT-B/32 CLIP        & 0.40 & 0.24 & 0.63 & 0.17 & 0.14 & 0.35 \\ 
 & ViT-B/32 $CF_{0.5}$  & \textbf{0.28} & \textbf{0.15} & \textbf{0.46} & \textbf{0.11} & \textbf{0.07} & \textbf{0.20} \\ 
 & ViT-B/16 CLIP        & 0.41 & 0.26 & 0.66 & 0.17 & 0.15 & 0.36 \\ 
 & ViT-B/16 $CF_{0.5}$ & \textbf{0.25} & \textbf{0.13} & \textbf{0.41} & \textbf{0.10} & \textbf{0.05} & \textbf{0.18} \\ 
\hline
\multirow{4}{*}{PATA} 
 & ViT-B/32 CLIP        & 0.55 & \textbf{0.19} & 0.81 & 0.56 & 0.23 & \textbf{0.93} \\ 
 & ViT-B/32 $CF_{0.5}$ & \textbf{0.41} & \textbf{0.19} & \textbf{0.70} & \textbf{0.55} & \textbf{0.20} & \textbf{0.93} \\ 
 & ViT-B/16 CLIP        & 0.46 & 0.18 & 0.78 & 0.55 & 0.23 & 0.95 \\ 
 & ViT-B/16 $CF_{0.5}$ & \textbf{0.39} & \textbf{0.13} & \textbf{0.63} & \textbf{0.55} & \textbf{0.18} & \textbf{0.90} \\ 

\bottomrule
\end{tabular}
\label{tab:bias_metrics}
\vspace{-1.0em}
\end{table*}

\begin{table*}[t!]
\centering
\footnotesize
\caption{Comparison of performance on downstream tasks for models fine-tuned on our synthetic dataset using weight ensembling set at $\alpha=0.5$. }
\setlength{\tabcolsep}{3pt}
\begin{tabular}{@{}lccc@{}}
\toprule
\textbf{Model} & \textit{FLICKR$_{R@5}$} & \textit{ImageNet1K$_{acc}$} \\
\midrule
ViT B/32 CLIP         & \textbf{0.83} & \textbf{0.63} \\
ViT B/32 $CF_{0.5}$   & 0.82 & 0.55 \\
ViT B/16 CLIP         & \textbf{0.86} & \textbf{0.68} \\
ViT B/16 $CF_{0.5}$   & \textbf{0.86} & 0.61 \\
\bottomrule
\end{tabular}
\label{tab:performance_metrics}
\vspace{-1.0em}
\end{table*}

\begin{table}[ht]
\centering
\caption{Recall of the image retrieval for subgroups on the PATA dataset. Using the ground truth sentences in the PATA dataset. Our method significantly improves the recall by several percentage points across all groups. H=Hispanic, I=Indian, E=Eastasian, B=Black, C=Caucasian; W=Woman, M=Man. These are the annotations provided in this dataset.}
\begin{tabular}{l|c|c|c|c|c|c|c|c|c|c||c}
\hline
Model & HW & HM & IW & IM & EW & EM & BW & BM & CW & CM & Accuracy\\
\hline
CLIP             & 0.86 & 0.86 & 0.82 & 0.85 & 0.84 & 0.84 & 0.87 & 0.82 & 0.87 & 0.90 & 0.81\\
$CF_{0.5}$       & \textbf{0.89} & \textbf{0.92} & \textbf{0.87} & \textbf{0.89} & \textbf{0.88} & \textbf{0.89} & \textbf{0.91} & \textbf{0.86} & \textbf{0.90} & \textbf{0.92} & \textbf{0.85} \\
\hline
$\Delta$ & \textcolor{ForestGreen}{+0.03} & \textcolor{ForestGreen}{+0.06} & \textcolor{ForestGreen}{+0.05} & \textcolor{ForestGreen}{+0.04} & \textcolor{ForestGreen}{+0.04} & \textcolor{ForestGreen}{+0.05} & \textcolor{ForestGreen}{+0.04} & \textcolor{ForestGreen}{+0.04} & \textcolor{ForestGreen}{+0.03} & \textcolor{ForestGreen}{+0.02} & \textcolor{ForestGreen}{+0.04}\\
\hline
\end{tabular}
\label{tab:pergroup_recall}
\end{table}

\subsection{Downstream Accuracy vs. Fairness Tradeoff}
\label{sec:results_alpha_blend}

\begin{figure}[tb]
  \centering
  \includegraphics[width=\textwidth]{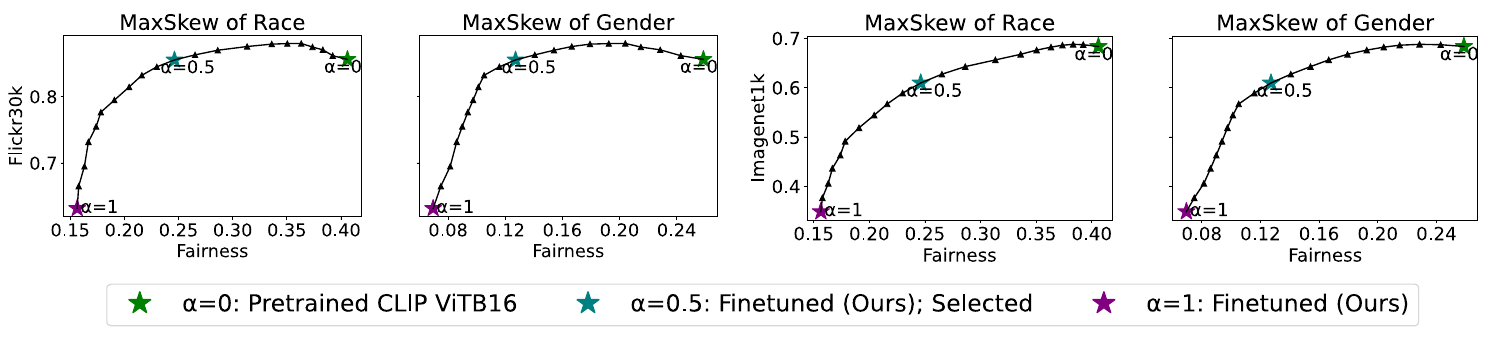}
  \caption{ The fairness and accuracy tradeoff as we vary $\alpha$ in weight space ensembling. Fairness is measured as the $MaxSkew@1k$ for FairFace and accuracy is measured with the Flickr30k (left) and Imagenet1k (right) datasets. }
  \label{fig:fairness_accuracy_tradeoff}
\end{figure}

Figure \ref{fig:fairness_accuracy_tradeoff} shows the accuracy-fairness tradeoff. An ideal tradeoff is a fairness metric closer to zero and higher downstream accuracy which corresponds to the top left of each subplot in Figure \ref{fig:fairness_accuracy_tradeoff}. Each marker corresponds to a different value of $\alpha$ as described in section \ref{sec:accuracy_fairness_tradeoffs}. We consistently note that as we increase $\alpha$, the fairness improves. Not only can we improve the fairness, but we can also improve the downstream accuracy, despite not fine-tuning on the downstream task. This behavior is exhibited in the subplots corresponding to Flickr30k, when $\alpha$ varies from 0.05 to 0.45, and for ImageNet1k, when $\alpha$ varies from 0 to 0.2 (top right).  

\subsection{Ablation Study}

\begin{table}[htbp]
\centering
\caption{Ablation experiments. Min/Max Skew @1k are reported with respect to race (R) and gender(G) on FairFace. All models use $\alpha=0.5$.}
\begin{tabular}{|l|l|l|l|l|}
\hline
\multicolumn{5}{|c|}{\textbf{Ablation on the number of views per caption.}} \\
\hline
 Counterfactuals & \textbf{$MaxSkew_R$} & \textbf{$MinSkew_R$}  & \textbf{$MaxSkew_G$} & $MinSkew_G$ \\
\hline
\(m=2\) & 0.26 & -0.44 & 0.15 & -0.19 \\
\(m=4\) & \textbf{0.24} & \textbf{-0.37 }&\textbf{0.12} & \textbf{-0.16} \\
\hline
\multicolumn{5}{|c|}{\textbf{Ablation on the dataset size.}} \\
\hline
Size & \textbf{$MaxSkew_R$} & \textbf{$MinSkew_R$}  & \textbf{$MaxSkew_G$}& \textbf{$MinSkew_G$} \\
\hline
25k & 0.30 & -0.45 & 0.17 & -0.24 \\
50k & 0.29 & -0.45 & 0.17 & -0.23 \\
75k & 0.27 & -0.42 & 0.17 & -0.22 \\
100k & \textbf{0.25} & \textbf{-0.40 }& \textbf{0.13} & \textbf{-0.16} \\
\hline
 
\end{tabular}
\label{tab:combined_ablation_experiments}
\end{table}

\begin{table}[ht]
\centering
\caption{Ablation on the weights of the combined loss function. $\beta_0$ is the weight for self supervised loss while $\beta_1$ is the weight for the original CLIP loss. Max Skew @24 is reported on PATA. The first row is equivalent to training with CLIP loss only.}

\begin{tabular}{|c|c|c|c|}
\hline
$\beta_0$ & $\beta_1$ &   \textbf{$MaxSkew_R$}  & \textbf{$MinSkew_R$} \\ \hline
        0     &     1       &        0.41           &    -0.75             \\ \hline
        \hline
        \hline
        0.25  &     1       &     0.42            &  -0.78                 \\ \hline
        0.50  &     1       &     0.41            &    	-0.86                  \\ \hline
        
        1     &     0.25   &      0.41           &      	-0.87              \\ \hline
        1     &     0.50   &     \textbf{0.39}	           &   -0.82                \\ \hline
        
         \hline
         \hline
        1     &      1      &   \textbf{0.39 }           &    	\textbf{-0.72 }              \\ \hline
        
\end{tabular}

\label{table:lossablation}
\end{table}

We study the effect of the the number of counterfactuals ($m$), the dataset size, and the loss weights. Tables \ref{tab:combined_ablation_experiments} and \ref{table:lossablation} show a summary of these results.

\noindent\textbf{The number of counterfactuals}, $m$, dictates how many counterfactuals will be used for each text caption in the batch. Table \ref{tab:combined_ablation_experiments} shows that doubling the number of counterfactuals contributes to better fairness, by bringing both skew metrics closer to zero.  To study, the \textbf{effect of the dataset size}, we first keep the size of the non-professions data the same ($\approx 100k$) and only change the size of the professions data. This is to mimic real life scenarios where the task specific data is usually limited. We find that as we increase the dataset size, fairness improves. This is due to the model being exposed to more diverse base images and their corresponding counterfactuals. To study the \textbf{effect of the loss}, we ablate the addition of the counterfactual loss introduced in Equation~\eqref{eq:lossfunction}. From Equation \eqref{eq:lossfunction}, \(\beta_0\) and \(\beta_1\) are weighting factors for the counterfactual and the original CLIP loss, respectively. In this experiment, we investigate how different weights affect the results, specifically, if the counterfactual loss brought any added benefits that the CLIP loss would not ensure. Thus by setting $\beta_0=0$ we would be finetuning with just the CLIP loss. Between the top row and bottom row, we find evidence to suggest that the counterfactual and CLIP loss together in equal proportion contribute to the best fairness. This is in inline with  \cite{mu2022slip} and \cite{tian2024stablerep}. The counterfactual loss only requires a few parameters for a projection layer and a  pairwise contrastive loss. When the counterfactual loss contributes more than the CLIP loss we have worse fairness (MinSkew). This is likely due to the fact that the CLIP loss is a function of the text and image embeddings, while the counterfactual loss is a function of the image embeddings only.  Consequently, the text encoder's parameters are updated less when the counterfactual loss dominates the overall loss.

\section{Conclusion and Limitations}
In this work, we introduced and validated a bias mitigation framework based on generating an entirely synthetic counterfactual dataset. The synthetic nature of our method not only enhances the privacy aspects of our work by circumventing the need for real photographs of individuals, but also ensures a high degree of data diversity. Our framework is generalizable to any concept we seek to debias, and is not limited to professions. 

\subsubsection{Limitations.} Synthetic counterfactuals have their own sets of biases stemming from the LLM and T2I models. We make simple modifications to address these stacked biases, such as neutralizing captions and negative prompting. Future work should investigate interpolating across diverse characteristics, like facial hair or skin tone, to generate more equitable and representative images. Our primary goal is to ensure that the inpainted regions offer sufficient contrast, supporting the generation of diverse counterfactuals necessary for robust contrastive learning.

\newpage

\bibliographystyle{splncs04}
\bibliography{egbib}
\end{document}


\appendix


\title{Supplementary Material} 
\author{Salma Abdel Magid\inst{1}\and
Jui-Hsien Wang\inst{2} \and
Kushal Kafle\inst{2}\and Hanspeter Pfister\inst{1}}
  
\institute{Harvard University \and
Adobe Research
}
\authorrunning{S.~Abdel Magid et al.}
\titlerunning{Supplementary Material}

\maketitle

\begin{figure}[H]

  \centering
  
  \includegraphics[width=\textwidth]{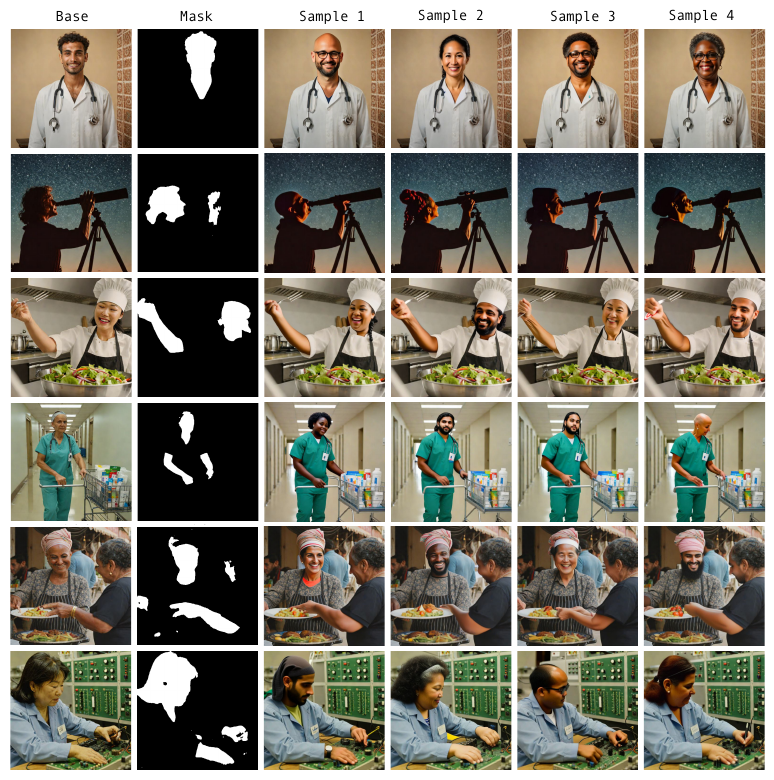}

  \caption{Sample images from our proposed fully 
  synthetic, fair, and private dataset.}
  \label{fig:suppgridimages1}
\end{figure}

\begin{figure}[H]
\includegraphics[width=\textwidth]{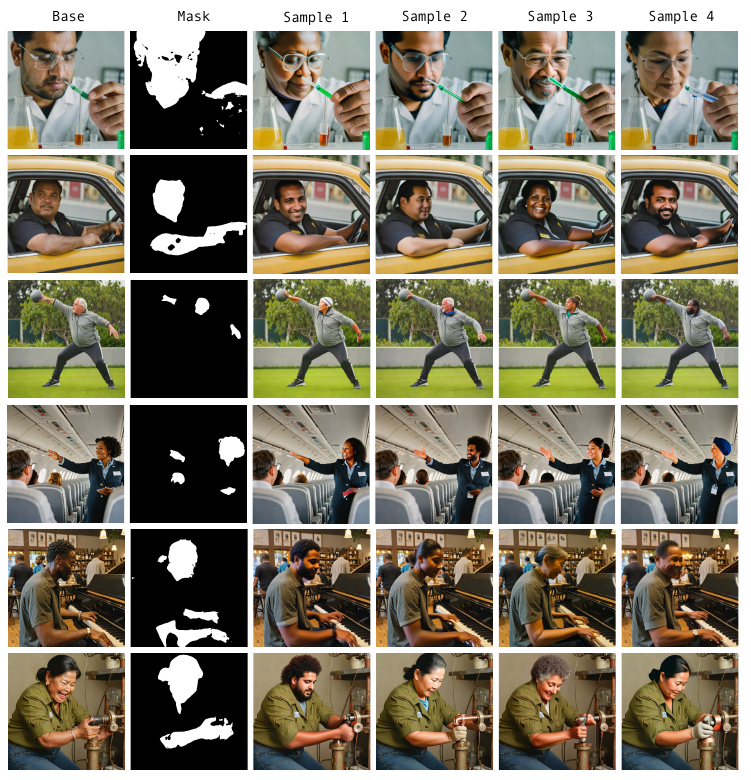}
  \caption{Sample images from our proposed fully 
  synthetic, fair, and private dataset.}
  \label{fig:suppgridimages2}
\end{figure}

\section{Extended Discussions On The Ethical Considerations Of Sensitive Attributes Categorization}
We acknowledge in the main text that sensitive attributes such as gender, race, and nationality are social constructs that cannot be captured in fixed categories. However, such constructs are still being used in two places of our pipeline. In this section, we explain the reasoning of this choice and provide more detailed analysis on its implications and limitations.

\paragraph{Racial and gender categorization.}
Usage of racial and gendered categorization schemes occurs at two points in our proposed pipeline: during dataset generation and during model evaluation. In the former, we use decorators for generating the base and inpainted images (\S\ref{suppmat:full_decorators}). In the latter, race and gender are directly used for evaluating the fairness of our fine-tuned model (via FairFace and PATA, as their annotations of sensitive attributes are crowdsourced). We discuss the limitations of this conceptualization below. 

\paragraph{Synthesizing diverse visuals using sampled decorators.} Our methodology for generating counterfactuals involves prompting a generative model with binary terms such as "man" and "woman". This approach can unintentionally reinforce the biases embedded in the model due to its specific architecture, objective, and training dataset. For instance, when generating images prompted with "woman", images often depict individuals wearing earrings despite negative prompting to avoid "jewellery". This phenomenon is evident in some of the examples presented in Figures~\ref{fig:suppgridimages1} and~\ref{fig:suppgridimages2}. 
Although this produces a minor artifact, it highlights that such biases can lead to unintended and misleading stereotypes.

However, it is crucial to note that our objective is not to teach CLIP what a man or woman looks like—indeed, CLIP is never exposed to captions that are not neutralized in our framework. Instead, our goal is to teach CLIP that, on average, what is relevant for identifying an attribute-neutral concept (e.g. profession) is precisely what lies outside the masked region. So long as the content within the masked region is sufficiently varied and in-distribution, its exact nature becomes increasingly irrelevant. This consideration leads us to question whether we can use masked or blurred images when retrieving images with CLIP.  Unfortunately, we cannot utilize masking or blurring since these methods would produce out-of-distribution images, constituting a completely different problem. Moreover, masking does not tackle the issue of secondary associations resulting from pretraining biases.

\paragraph{Model evaluation with a fixed taxonomy of sensitive attributes.} In a similar vein, it is important to recognize the constraints of current evaluation practices, particularly those relying on existing gender and race crowdsourced annotations. These methods fail to capture the full spectrum of human diversity. However, our work primarily introduces a novel approach for debiasing models, rather than focusing on evaluation metrics. One direct extension of our work can begin to address this by generating many more diverse counterfactuals, then grouping them with unsupervised clustering, similar to StableBias~\cite{luccioni2024stable}. The fairness groups are then based on observable characteristics—such as skin tone, facial features, or attire—rather than crowdsourced, socially constructed categories. 
Furthermore, efforts to balance representation across different groups in datasets encounters the complex issue of latent biases. Achieving numerical balance does not eliminate the risk of reinforcing stereotypes or establishing spurious correlations. For instance, a dataset, even if balanced in terms of race and gender, might inadvertently associate certain groups with specific backgrounds, settings, or emotions, thereby perpetuating existing biases. Our method can begin to address these issues during the development of evaluation datasets. 

\section{LLM-Generated Captions}

\paragraph{Prompt for LLM.}
\label{sec:llmprompt}

The exact prompt used for generating neutral captions using LLAMA 70B caption generation in our system is:

\begin{quote}
\emph{I need to generate a set of $\{m\}$ unique and very different descriptions, focusing on a specific profession. Use gender neutral words. The profession for this task is $\{P_i\}$. Based on a detailed description of this $\{P_i\}$, along with its types, synonyms, and key aspects, please create image captions that: \textcolor{purple}{1.} are a single sentence with a scenario (no names, just profession synonyms e.g. A $\{P_i\}$ carrying a hamburger, fries, and drink to a table) \textcolor{purple}{2.} depict synonyms or related roles of the $\{P_i\}$, such as types or specialization of $\{P_i\}$ but neutralize any gendered ones like busboy -> busperson, waitress -> waiter \textcolor{purple}{3.} depict the tools, equipment, and work environment typical for the $\{P_i\}$. Important! Ensure all descriptions are gender-neutral and avoid racial language, focusing instead on the activities and settings of the profession. No names. Do not to use names, or gendered language (such as boy, waitress, she, he).}
\end{quote}

Using this template, we generate a set of captions for all the professions. Table~\ref{tab:suppmatprofessions} includes all 278 professions we used in this work. We note that the specific LLM prompt outlined in this section will need to be modified to extend this to other visual concepts that are not professions.

\paragraph{Examples of LLM-generated captions.}

Table \ref{table:suppmat_sample_captions} shows a set of sample captions generated using our framework. These captions are then used for two purposes: (1) to generate the base and inpainted images and (2) as the ground truth caption in the training dataset.

\begin{table}
\tiny
\centering
\caption{\small The set of professions used. Please zoom in.}
\label{tab:suppmatprofessions}
\begin{tabular}{|l|l|l|l|}
\hline
academic & accountant & actor & advertising agent \\
\hline
aerospace engineer & aide & air conditioning installer & air traffic controller \\
\hline
aircraft cargo handler & aircraft pilot & ambulance driver & analyst \\
\hline
announcer & architect & artist & artist agent \\
\hline
astronomer & athlete & attendant & audiologist \\
\hline
auditor & author & baggage porter & bailiff \\
\hline
baker & banker & barber & bartender \\
\hline
biologist & bookkeeper & boxer & broker \\
\hline
bus driver & businessperson & butcher & car mechanic \\
\hline
career counselor & carpenter & carpet installer & cashier \\
\hline
ceo & chef & chief & chief executive \\
\hline
childcare worker & chiropractor & choreographer & civil engineer \\
\hline
claims appraiser & cleaner & cleaning person & clergy \\
\hline
clerk & coach & comedian & commercial and industrial designer \\
\hline
community manager & compliance officer & composer & computer programmer \\
\hline
computer support specialist & computer systems analyst & construction laborer & construction worker \\
\hline
cook & correction officer & counselor & courier \\
\hline
credit analyst & credit counselor & criminal investigator & curator \\
\hline
customer service representative & dancer & data entry keyer & dental assistant \\
\hline
dental hygienist & dentist & designer & detective \\
\hline
dietitian & diplomat & dishwasher & dispatcher \\
\hline
doctor & drawer & dry-cleaning worker & drywall installer \\
\hline
economist & editor & electrical engineer & electrician \\
\hline
elementary school teacher & embalmer & engineer & environmental engineer \\
\hline
event planner & executive assistant & explosives worker & facilities manager \\
\hline
farmer & farmworker & fast food worker & file clerk \\
\hline
financial advisor & fine artist & firefighter & fisher \\
\hline
fitness instructor & flight attendant & food preparation worker & fundraiser \\
\hline
glazier & graphic designer & groundskeeper & guard \\
\hline
hairdresser & handball player & handyperson & hazardous materials removal worker \\
\hline
head cook & health technician & healthcare social worker & highway maintenance worker \\
\hline
host & hostess & housekeeper & illustrator \\
\hline
industrial engineer & insurance agent & insurance sales agent & interior designer \\
\hline
interviewer & inventory clerk & it specialist & jailer \\
\hline
janitor & jeweler & journalist & judge \\
\hline
kindergarten teacher & laboratory technician & laborer & language pathologist \\
\hline
lawyer & legislator & librarian & linguist \\
\hline
lodging manager & logging worker & logistician & machinery mechanic \\
\hline
machinist & maid & maintenance worker & manager \\
\hline
manicurist & market research analyst & marketing manager & massage therapist \\
\hline
material mover & mathematician & mechanic & mechanical engineer \\
\hline
medical records specialist & medical scientist & mental health counselor & metal worker \\
\hline
meter reader & midwife & model & mortician \\
\hline
motion picture projectionist & mover & movie director & musician \\
\hline
network administrator & news analyst & nuclear engineer & nurse \\
\hline
nursing assistant & nutritionist & occupational therapist & office clerk \\
\hline
office worker & opera singer & optician & painter \\
\hline
paralegal & paramedic & parking lot attendant & payroll clerk \\
\hline
pensioner & personal care aide & pest control worker & pharmacist \\
\hline
pharmacy technician & photographer & physical therapist & physician \\
\hline
physicist & pianist & pilot & plane mechanic \\
\hline
plasterer & plumber & podiatrist & poet \\
\hline
police officer & postal service worker & postal worker & presser \\
\hline
priest & printing press operator & producer & professor \\
\hline
programmer & proofreader & psychologist & public relations specialist \\
\hline
purchasing agent & radio operator & radiologic technician & rancher \\
\hline
real estate broker & receptionist & recreation worker & repair worker \\
\hline
reporter & retail salesperson & roofer & sailor \\
\hline
sales manager & salesperson & school bus driver & scientist \\
\hline
secondary school teacher & secretary & security guard & sheet metal worker \\
\hline
sheriff & ship captain & shoe worker & singer \\
\hline
social worker & sociologist & software developer & soldier \\
\hline
special education teacher & statistician & stocker & subway operator \\
\hline
supervisor & surgeon & survey researcher & swimmer \\
\hline
tailor & tax examiner & taxi driver & teacher \\
\hline
teaching assistant & telemarketer & telephone operator & teller \\
\hline
tennis player & theologian & therapist & tour guide \\
\hline
tractor operator & travel agent & truck driver & tutor \\
\hline
umpire & underwriter & veterinarian & waiter \\
\hline
welder & wholesale buyer & & \\ 
\hline

\end{tabular}

\end{table}

\begin{table}[!ht]
\centering
\caption{Sample generated captions for various professions.}
\label{table:suppmat_sample_captions}
\begin{adjustbox}{margin=0em -1em 0em 0em}  
\scriptsize
\begin{tabular}{|l|p{8cm}|}   

\hline
\textbf{Profession} & \textbf{Sample Caption} \\
\hline
surgeon & \parbox{8cm}{\raggedright gastroenterologist performing an endoscopy to examine a patients digestive tract} \\\hline
facilities manager & \parbox{8cm}{\raggedright manager supervising the installation of new security cameras in a parking lot} \\\hline
mechanical engineer & \parbox{8cm}{\raggedright a close-up shot of someones hands carefully aligning gears in a mechanical transmission system} \\\hline
dentist & \parbox{8cm}{\raggedright close-up shot of a dentists hands, with a set of forceps and a dental implant in hand, placing the implant into a patients jawbone} \\\hline
advertising agent & \parbox{8cm}{\raggedright advertising agent sitting in a conference room, nodding their head and taking notes during a meeting} \\\hline
painter & \parbox{8cm}{\raggedright painter in a gallery, carefully examining the texture and color of a finished piece} \\\hline
environmental engineer & \parbox{8cm}{\raggedright environmental engineer assessing the air quality in a subway tunnel using a portable monitor} \\\hline
plasterer & \parbox{8cm}{\raggedright plasterer wearing protective eyewear and gloves, using a sanding block to smooth out a freshly plastered wall} \\\hline
graphic designer & \parbox{8cm}{\raggedright graphic designer in a home office, using a large monitor and keyboard to create a social media campaign for a client} \\\hline
childcare worker & \parbox{8cm}{\raggedright childcare worker workers in a music room, playing instruments and singing songs with a group of preschoolers} \\\hline
chef & \parbox{8cm}{\raggedright close-up shot of a chefs hands as they chop fresh herbs on a cutting board} \\\hline
healthcare social worker & \parbox{8cm}{\raggedright A healthcare professional wearing a stethoscope around their neck, standing in front of a medical supply cart} \\\hline
fitness instructor & \parbox{8cm}{\raggedright fitness instructor Personal trainer correcting a clients form during a squat exercise in a private training session} \\\hline
air conditioning installer & \parbox{8cm}{\raggedright air conditioning installer working on an air conditioning system in a cleanroom environment} \\\hline
tailor & \parbox{8cm}{\raggedright tailor holding a pair of pants up to a customers waist, marking the hemline with chalk} \\\hline
tour guide & \parbox{8cm}{\raggedright tour guide wearing a waterproof jacket and carrying a fishing rod stands on the shore of a lake, demonstrating a casting technique to a group of anglers} \\\hline
handball player & \parbox{8cm}{\raggedright close-up shot of a handball players fingers, showing their dexterity and grip on the ball} \\\hline
radiologic technician & \parbox{8cm}{\raggedright radiologic technician maintaining accurate and detailed records of radiological exams and procedures} \\\hline
fundraiser & \parbox{8cm}{\raggedright fundraiser holding a Thank You sign in front of a donation box} \\\hline
professor & \parbox{8cm}{\raggedright professor holding a petri dish, examining the growth of bacteria on an agar plate} \\\hline
nuclear engineer & \parbox{8cm}{\raggedright nuclear engineer conducting a safety training drill for nuclear power plant employees} \\\hline
choreographer & \parbox{8cm}{\raggedright a large gymnasium, a choreographer stands on a balance beam, demonstrating a series of acrobatic movements to a group of dancers} \\\hline
physical therapist & \parbox{8cm}{\raggedright physical therapist sitting in a warm water pool, performing arm circles with their hands submerged in the water} \\\hline
curator & \parbox{8cm}{\raggedright curator standing on a ladder, hanging a large piece of artwork on a wall in a gallery} \\\hline
legislator & \parbox{8cm}{\raggedright legislator speaking on the phone, possibly with a headset, in a quiet office or workspace} \\\hline

\hline

\end{tabular}
\end{adjustbox}
\end{table}

\section{Additional Details: Generating Base Images}
\subsection{Full Decorators}
\label{suppmat:full_decorators}
Table \ref{table:supp_decorator_prompts} presents the comprehensive list of decorators utilized in our image generation process. The format for generating image instructions is as follows: 
\\
A \texttt{[shot\_style]} photo of a \texttt{[body\_type]}, \texttt{[age]}, \texttt{[skin\_color]}, \texttt{[nationality]} \texttt{[gender]} \texttt{[caption]}. This instruction is then appended with "\texttt{with no hair, bald}" or "\texttt{with [hair\_length],[hair\_color], [hair\_style],[hair\_texture] hair}" depending on the hair style decorator value.

\begin{table}[ht]
\centering
\caption{Full list of decorators for image generation.}
\label{table:supp_decorator_prompts}
\begin{adjustbox}{width=\columnwidth,totalheight=\textheight,keepaspectratio,center}

\begin{tabular}{ll}
\hline
\textbf{Attribute} & \textbf{Examples} \\
\hline
Shot Style & iPhone, Long shot, Upper body shot, Stock \\
Nationality & Nigerian, Sudanese, Moroccan, Egyptian, Kenyan, Afghan, Indian, Iranian, \\
& Pakistani, Syrian, Turkish, Italian, German, French, Mexican, Bolivian, Brazilian, \\
& Guatemalan, Saudi, Japanese, Vietnamese, Chinese, Indonesian, Korean
\\
Age & Young (20s), Middle-aged (30s-40s), Mature (50s+) \\
Gender & Woman, Man \\
Skin Color & Brown, Black, Light \\
Body Type & Fat, Skinny, Chubby, Athletic \\
Hair Color & Black, Brown, Red, Grey \\
Hair Lengths & Short, Long \\
Hair Textures & Straight, Curly, Wavy \\
Hair Style & Afro, Long, Bald, Ponytail, Braids \\
\hline
\end{tabular}
\end{adjustbox}
\end{table}

\subsection{Additional Examples of Counterfactuals}
Figures~\ref{fig:suppgridimages1} and~\ref{fig:suppgridimages2} show additional examples of the counterfactuals generated using our framework. These are then used for finetuning CLIP with neutralized captions, such as those in Table \ref{table:suppmat_sample_captions}. 

\section{Examples of Failure Cases}
\begin{figure}[H]
\includegraphics[width=\textwidth]{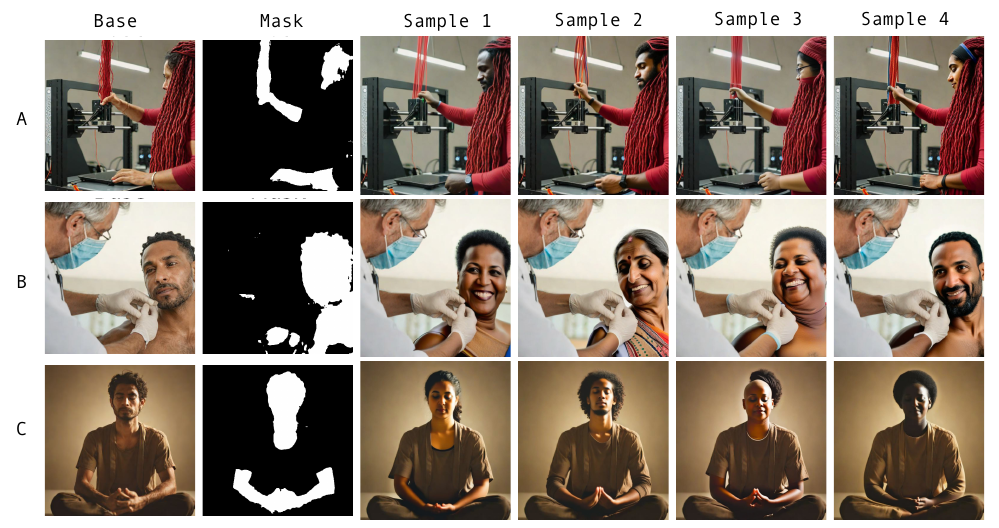}
  \caption{Examples of failure cases.}
  \label{fig:suppfailures}
\end{figure}
Figure~\ref{fig:suppfailures} shows examples of failure cases. We consider row A and B in Figure~\ref{fig:suppfailures} to be two types of mask failures. Row A demonstrates that the individual's hair is not masked correctly, and thus is present in all the generated counterfactuals.  Row B demonstrates that the doctor was not masked at all, but the patient was. The caption for row B is "doctor administering a vaccine to a patient." One way to address this masking issue is to further modify the prompt for the caption generation  (\S~\ref{sec:llmprompt}) so that it only generates captions with a single individual.

Lastly, row C has two issues. The neck in sample 1 is more exposed compared to the other samples in this generation. Although the region is correctly masked, the generation is not the same across samples. Moreover, the instruction for sample 3 is: "A photo of a fat, Young (20s), Black, Saudi, woman with no hair, bald therapist sitting in a meditation position, eyes closed, surrounded by soft lighting and peaceful music." However, the image clearly contains some hair, despite the "with no hair, bald" instruction.

\section{Fairness Metrics Definitions}
We quantify bias in our model using standard metrics including
maximum/minimum skew, and the normalized discounted Kullback-Leibler divergence (NDKL). Skew captures the largest unfair advantage or disadvantage across attributes, while NDKL is a retrieval metric that measures divergence from an unbiased distribution. Specifically, given a ranked list of images corresponding to a text query $\tau_q$, let $p^{k}_{a}$ be the proportion of the top $k$ images annotated with the attribute $a$. Then,  the skew of attribute $a_{i}$ in the top $k$ is $skew_{a_{i}} @ k =log \frac{p_{a_{i}}^k}{1/|A|}$ where $A$ is set of attribute values. To measure the largest unfair advantage and worst disadvantage for any individuals with a specific attribute value, we measure, respectively,
 
\begin{itemize}
    \item MaxSkew: $MaxSkew@k = max_{a_{i}\in A} skew_{a_{i}} @ k$
    \item MinSkew: $MinSkew@k = min_{a_{i}\in A} skew_{a_{i}} @ k$.
\end{itemize} 

The skew-based metrics only capture the skew of a single attribute. To address this, the NDKL operates over \textit{all} attribute values by computing a normalized KL divergence between the distribution of attributes in the top-$k$ and a desired distribution. In a fair setting,  the desired distribution should be the uniform distribution of attributes, that is the proportion of all attributes should be equal in the top-$k$ retrieved images. For a text query $\tau_q$, $\text{NDKL}(\tau_q)=\frac{1}{Z} \sum^{|\tau_q|}_{i=1} \frac{1}{log_{2}(i+1)} d_{KL}(D_{\tau_q}||D_u)$ where, $d_{KL}(\cdot)$ is the KL divergence, $Z$ is a normalization constant equal to $Z=\sum^{|\tau_q|}_{i=1} \frac{1}{log_{2}(i+1)}$, and $D_u$ is the uniform distribution of attributes.

\bibliographystyle{splncs04}
\bibliography{egbib}